%% file: InterpretationOfFCLayer.tex
\documentclass[twoside,twocolumn]{article}

\usepackage[sc]{mathpazo} 
\usepackage[T1]{fontenc} 
\usepackage{microtype} 

\usepackage[english]{babel} 

\usepackage[hmarginratio=1:1,top=32mm,columnsep=20pt]{geometry} 
\usepackage[hang, small,labelfont=bf,up,textfont=it,up]{caption} 
\usepackage{booktabs} 

\usepackage{abstract}

\usepackage{titlesec} 
\renewcommand\thesection{\Roman{section}} 
\renewcommand\thesubsection{\roman{subsection}} 
\titleformat{\section}[block]{\large\scshape\centering}{\thesection.}{1em}{} 
\titleformat{\subsection}[block]{\large}{\thesubsection.}{1em}{} 

\usepackage{fancyhdr} 
\usepackage{titling} % Customizing the title section

\usepackage{hyperref} % For hyperlinks in the PDF
\usepackage{amsmath,stackengine}

\usepackage{algorithm}
\usepackage{algpseudocode}

\usepackage{graphicx} % more modern
\usepackage{caption}
\usepackage{subcaption}
\usepackage{wrapfig}

\usepackage{tikz}
\include{defs}

\pretitle{\begin{center}\Huge\bfseries} % Article title formatting
\posttitle{\end{center}} % Article title closing formatting
\title{An interpretation of the final fully connected layer} % Article title
\author{%
\textsc{Siddhartha} \\[1ex]
\normalsize Saarland University\\ 
}
\date{}
\renewcommand{\maketitlehookd}{%
\begin{abstract}
\noindent 
In recent years neural networks have achieved state of the art accuracy for various tasks 
but the explainability of the generated outputs still remains difficult. In this work we
attempt to provide a method to understand the learnt weights in the final fully connected layer in
image classification models. We motivate our method by drawing a connection
between the policy gradient objective in reinforcement learning and the supervised learning objective. We suggest
that the commonly used cross entropy based supervised learning objective can be regarded as a 
special case of the policy gradient objective. Using this insight we propose a method to find the most discriminative and
confusing parts of an image. Our method does not make any prior assumption about neural network achitecture and has
low computational cost. We apply our method on publicly available pre-trained models and report the generated results.
\end{abstract}
}

\begin{document}
\usetikzlibrary{trees,arrows}

% Print the title
\maketitle

\section{Introduction}

The two most popular paradigms of deep neural network training namely, Reinforcement learning and supervised learning attempt to address different problems. In supervised learning 
the aim is to learn a function that best approximates a training dataset with the hope that the learnt model weights generalize on an unseen test set. Thus, the training set data
plays a major role in the performance of the trained model. In many real world tasks such well-labelled datasets may be available in a limited amount and difficult to generate. Even if
a well-labelled big dataset is available, the supervised learning approach cannot handle a a distribution shift during test time. Since the learnt model weights are static, it may be difficult
to apply such algorithms in dynamic environments. 
Reinforcement learning can be regarded as an attempt to address the limitations of the supervised learning approach. Different RL methods can again be broadly classified
into two categories depend on how the policy is learnt. In online reinforcement learning, the aim is to let the agent interact with the environment and use the collected data to 
optimize the reward values. In contrast, offline reinforcement learning attempts to learn a policy from an existing dataset. Considering the lack of interaction with the environment
and the fixed policy, similar to supervised learning, offline RL faces the challenge of distribution shift. The impact of distribution shift on offline RL is of even more signifance in the RL
context since the whole purpose of training the agent was online interaction with an environment. Methods like off-policy policy gradients and approximate off-policy policy gradients
try to address this shortcoming. We refer the reader to \cite{levine2020offline} for a survey of different offline RL methods and 
their limitations. With regards to this work, offline policy gradient method is of primary concern. So we provide a brief overview of the offline policy gradient method.
\subsection*{Offline Policy Gradient Method}
Let $\bs_t$, $\ba_t$ and $r(\bs_t,\ba_t)$ denote the state, action and reward of an agent at time $t$. 
Let $\policy_\theta$ denote the agent policy parameterised by $\theta$. $\traj$ denotes the trajectory obtained by 
following the policy $\policy_\theta$. It is a tuple of state and action values,
$\traj = (s_0,a_0,s_1,a_1,..s_T,a_T)$. The trajectory distribution $p_{\policy_\theta}(\traj)$ under the policy ${\policy_\theta}$ can be expressed 
as $p(s_1)\prod_{t=0}^T{\policy_\theta}p(\bs_{t+1}|\bs_t, \ba_t)$
The reinforcement learning objective, $J(\policy_\theta)$, can then be expressed as an expectation under this trajectory distribution:
\begin{equation}
J({\policy_\theta}) = \E_{\traj \sim p_{\policy_\theta}(\traj)}\left[
\sum_{t=0}^T \reward(\bs_t, \ba_t)
\right]. \label{eq:rl_objective}
\end{equation}
The goal of RL is to learn the optimal parameter $\theta^*$ such that equation \ref*{eq:rl_objective} is maximized. 
\begin{equation}
    \theta^* = \arg\max_{\theta}\E_{\traj \sim p_{\policy_\theta}(\traj)}\left[
    \sum_{t=0}^T \reward(\bs_t, \ba_t)
    \right]. \label{eq:opt_theta}
\end{equation}
    
The gradient of the objective $J({\policy_\theta})$ with respect to $\theta$ is given by
\begin{equation}
    \nabla J({\policy_\theta}) = \E_{\traj \sim {\policy_\theta}}[(\sum_{t=1}^T \nabla log {\policy_\theta}(\ba_t,\bs_t))]
    (\sum_{t=1}^T r(\bs_t,\ba_t)] \label{eq:rl_grad_objective}
\end{equation}

We refer the reader to \cite{sutton1999policy} for a detailed derivation and discussion of the policy gradient method.
%------------------------------------------------

\section{Connecting Supervised learning with Policy Gradient Objective}
In this section we try to draw a connection between the current supervised learning paradigm and policy gradient objective. In particular, we make the claim
that SL is equivalent to offline policy gradient RL, the SL policy is a function of the dataset distribution instead of the model parameters.\\
In the supervised learning setting we attempt to learn a function parameterised by $\theta$ to minimize the expected loss $L(\hat{y}_\theta, y)$ between 
the true labels $y$ and predicted labels $\hat{y}$.
The vanilla SL objective $S(\theta)$ can then be stated as
\begin{equation}
    S(\theta) = \arg \min_{\theta} \E [L (\hat{y}_{\theta}, y)]
\end{equation}
If we set $R = - L (\hat{y}_{\theta}, y)$, we can rewrite the objective as 
\begin{equation}
    S(\theta) = \arg \max_{\theta} \E [R (\hat{y}_{\theta}, y)]
\end{equation}
The gradient of the objective with respect to $\theta$ is
\begin{equation}
    \nabla_{\theta} S(\theta) = \E [\nabla_{\theta} R (\hat{y}_{\theta}, y)]
    \label{eq:sl_grad_objective}
\end{equation}
When we compare the gradient of SL objective \ref{eq:sl_grad_objective}  with RL objective \ref{eq:rl_grad_objective}, we can notice the absence of the policy gradient term  $\nabla_{\theta} log {\policy_\theta}(\ba_t,\bs_t))$.\\
What is the policy for the supervised learning objective ? We address this concern by referring back to the RL objective \ref{eq:rl_objective}.
We notice that if $\nabla_{\theta} log {\policy_\theta}(\ba_t,\bs_t))$ is set to 1, then the RL objective \ref{eq:rl_objective} is identical to SL objective. 
We observe that if $\theta$ was drawn from an exponential family of distributions of the form $exp(a*\theta+b)$, where
$a$ and $b$ are some scalars then $\nabla_{\theta} log {\policy_\theta}(\ba_t,\bs_t)) = 1$. We propose that $\theta$ follows an Asymmetric Laplace Distribution (ALD) given by
\begin{equation}
    f(\theta, m, \lambda, \kappa) = \frac{\lambda}{\kappa+1/\kappa}
    \begin{cases}
        \exp((\lambda/\kappa)(\theta-m)) & (\theta < m)\\
        \exp((-\lambda\kappa)(\theta-m)) & (\theta >= m)
    \end{cases}
\end{equation}
$\lambda$ and $\kappa$ are constants dependant on the training dataset and $m$ is the location parameter.
We provide a sketch of our intuition here.\\
During model training using supervised training, we note that for each sample in a batch there are two possibilities, namely
\begin{itemize}
        \item The sample belongs to one of the classes in the training dataset.
        \item The sample belongs to some class that is outside the training dataset. \label{possib}
\end{itemize}
Both these possibilities are mutually exclusive and can be modelled as a stationary exponential distribution with corresponding rate parameters.
We suggest an exponential distribution because the selection of a sample from a particular class can be modeled by a Poisson distribution if the samples are assumed to be independent and distributed identically in the dataset.
The rate parameters for the exponential distribution will be determined by the class frequency in the training dataset. We hypothesize that the trained model 
weights are trying to maximize the difference between the two possibilities \ref{possib} for each class in the dataset. We couple this assumption with the fact that the difference of two RV drawn from an exponential distributions is
a Laplace distribution \cite{diff2exp}. This allows us to conclude that the model parameters should follow a Laplace distribution. This distribution will be assymetric because the rate 
parameter for both the outcomes \ref{possib} are different in the general case.
% Let us consider the class $k$. Let $\hat{y_k}$ denote the predicted probability for the $k_th$ class. This implies that the probability of the sample to be from 
% a class outside the training dataset is given by $1-\hat{y_k}$.
% $\hat{y_k} = \theta_1^TX$ and $1 - \hat{y_k} = \theta_2^TX$. Both $\hat{y_k}$ and $1-\hat{y_k}$ are exponentially distributed and mutually exclusive.
% Taking the difference of both quantities, we get $2*\hat{y_k} - 1 = (\theta_1 - \theta_2)^TX$.  Let $(\theta_1 - \theta_2) \sim f(\theta, m, \lambda, \kappa)$
% We take expectation with respect to $\theta$ on both sides which gives us $E_\theta [2*\hat{y_k} - 1] = E_\theta [(\theta_1 - \theta_2)^TX]$.
% We have already reasoned that $E_\theta[\hat{y_k}]$ is an exponential distribution. Also note that $X$ is independent of $\theta$.
% $2exp(a_1*x) - 1 = exp(a_2*(x-m) + b_2)$. This implies that $\theta_1$ and $\theta_2$ follow an asymmetric Laplace distribution with $m=0$
% We can interpret the above result to indicate that once the model training is completed, a subset of model weights are activated to indicate the 
% probability for kth class $\hat{y_k}$. The other subset determines the probability for the sample to be outside the training set. Since both these outcomes
% are mutually exclusive it makes intuitive sense that the weights modelling either scenario are opposite in sign \($m=0$\)

\begin{algorithm}[ht]
    \caption{Fitting FC layer weights with ALD \label{alg:fitALD}}
    \begin{algorithmic}[1]
    \State Let $\theta_k$ denote the final FC layer weight for $kth$ class
    \For{class $k \in [0, \dots, K]$}
    % \State initialize $m_0 = mean(\theta_k)$
    % \State $m \leftarrow mean(\theta_k) $ \Comment{Mean weight for class $k$}
    \State $\theta_{pos} \leftarrow \theta_k[\theta_k > 0]$ \Comment{Isolate the weights for the positive outcome}
    \State Let $L = len(\theta_{pos})$
    \State Initialize $x_0 = [0, 1/L, 2/L, \dots 1]$
    \State Minimise LSQ error on $|\log(\theta_{pos}) - (a*x_0 + b)|_2$. \Comment{Use Linear regression to estimate slope \& intercept}
    \EndFor
    \end{algorithmic}
\end{algorithm}
    
%------------------------------------------------

\section{Results}
We used our proposed interpretation of final fully connected layer weights to fit ALD distribution to the pre-trained weights of popular architectures available on timm \cite{rw2019timm}.
We attach plots comparing the actual weights with our ALD fit. Due to space constraints, we report results only for Swin \cite{liu2021swin}, Resnet 18 and Resnet152 \cite{he2016identity}. We tried to keep our
selected architectures to be diverse to show the wide applicability of our proposed fitting method. Resnet18 was selected as an example of a very small CNN based model, Resnet152 was selected to 
represent large CNN based models and Swin was chosen as a representative of the vision transformer class of models. For each architecture, we consider three imagenet classes, namely tricycle, web site \& whiptail.
The classes were randomly chosen and similar results can be verified if other classes are considered. It can be observed from the plots that for the line obtained from our linear regression 
is a good fit for a large subset of weights. Class wise plots for different architectures can be found at this \href{https://drive.google.com/file/d/1ce90RTQKhYIoxhJrqvhxw6VdjLMEGHkm/view?usp=sharing}{link}.
\begin{figure}
  \centering
  \begin{subfigure}{0.2\textwidth}
      \includegraphics[width=\textwidth]{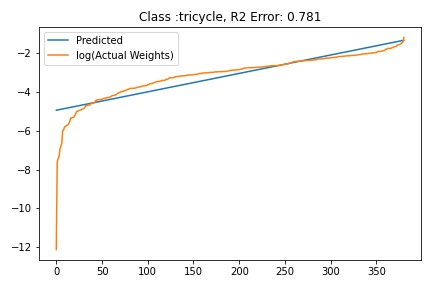}
      % \caption{Firts subfigure.}
      % \label{fig:first}
  \end{subfigure}
  \hspace{0.2mm}
  \begin{subfigure}{0.2\textwidth}
      \includegraphics[width=\textwidth]{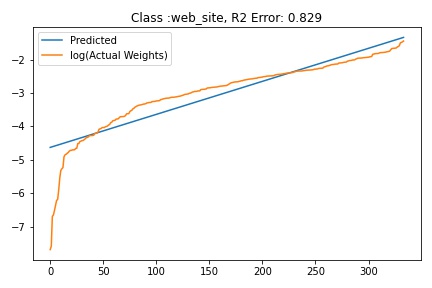}
      % \caption{Second subfigure.}
      % \label{fig:second}
  \end{subfigure}
  \hspace{0.2mm}
  \begin{subfigure}{0.2\textwidth}
      \includegraphics[width=\textwidth]{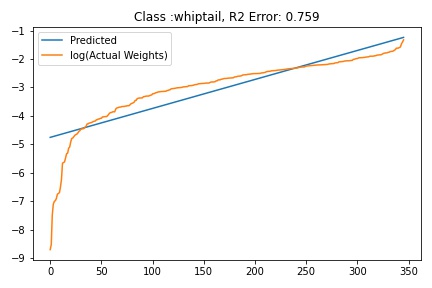}
      % \caption{Third subfigure.}
      % \label{fig:third}
  \end{subfigure}
          
  \caption{Fitting ALD to Resnet152.  Note that the FC layer dimension is $1000 \times 2048$.}
  \label{fig:resnet152_fit}
\end{figure}

\begin{figure}
  \centering
  \begin{subfigure}{0.2\textwidth}
      \includegraphics[width=\textwidth]{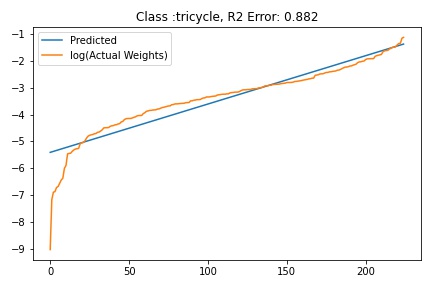}
      % \caption{Firts subfigure.}
      % \label{fig:first}
  \end{subfigure}
  \hspace{0.2mm}
  \begin{subfigure}{0.2\textwidth}
      \includegraphics[width=\textwidth]{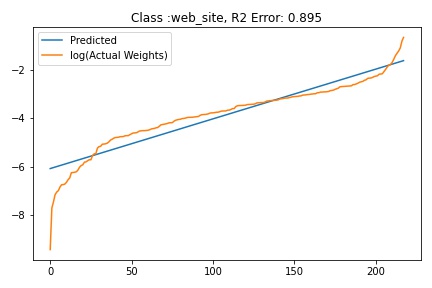}
      % \caption{Second subfigure.}
      % \label{fig:second}
  \end{subfigure}
  \hspace{0.2mm}
  \begin{subfigure}{0.2\textwidth}
      \includegraphics[width=\textwidth]{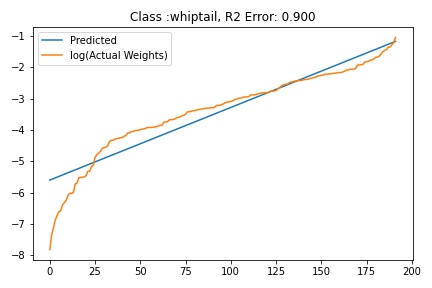}
      % \caption{Third subfigure.}
      % \label{fig:third}
  \end{subfigure}
          
  \caption{Fitting ALD to the FC layer of Resnet18. Note that the FC layer dimension is $1000 \times 512$.}
  \label{fig:resnet18_fit}
\end{figure}

\begin{figure}[h!]
  \centering
  \begin{subfigure}{0.2\textwidth}
      \includegraphics[width=\textwidth]{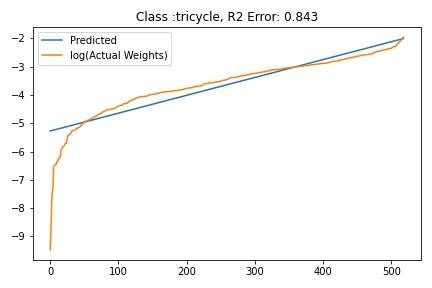}
  \end{subfigure}
  \hspace{0.2mm}
  \begin{subfigure}{0.2\textwidth}
      \includegraphics[width=\textwidth]{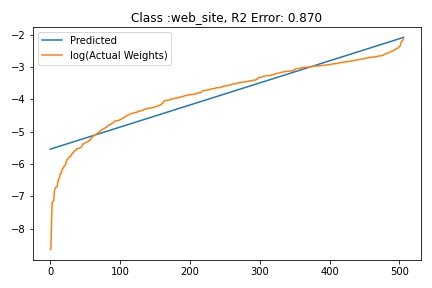}
  \end{subfigure}
  \hspace{0.2mm}
  \begin{subfigure}{0.2\textwidth}
      \includegraphics[width=\textwidth]{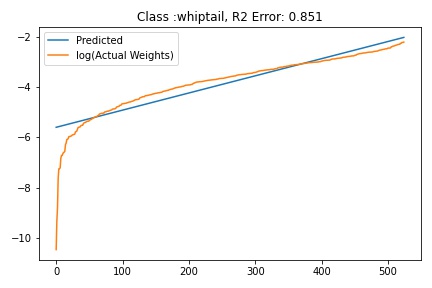}
  \end{subfigure}
          
  \caption{Fitting ALD to Swin Base transformer \cite{liu2021swin} trained with $384\times384$ image size.  Note that the FC layer dimension is $1000 \times 1024$.}
  \label{fig:swin_base}
\end{figure}

%------------------------------------------------
\section{Discussion}
% https://stackoverflow.com/questions/47388728/how-to-make-the-tree-like-graph-in-latex
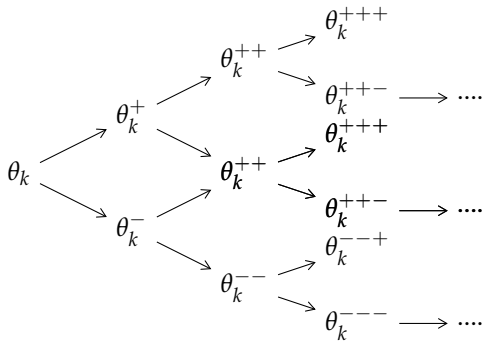
\begin{figure}[ht!]
    
  \tikzstyle{level 1}=[level distance=15mm, sibling distance=15mm]
  \tikzstyle{level 2}=[level distance=15mm, sibling distance=15mm]
  \tikzstyle{level 3}=[level distance=15mm, sibling distance=10mm]
  \begin{tikzpicture}[grow=right,->,>=angle 60]
    \node {$\theta_k$}
      child {node {$\theta_k^{-}$}
        child {node {$\theta_k^{--}$}
          child{node{$\theta_{k}^{---}$}
          child{node{$....$}}
          }  
          child{node{$\theta_k^{--+}$}}
        }
        child {node{$\theta_k^{-+}$}
          child{node{$\theta_{k}^{-+-}$}
          child{node{$....$}}
          }  
          child{node{$\theta_k^{-++}$}}
        }
      }
      child {node {$\theta_k^{+}$}
        child {node{$\theta_k^{+-}$}
          child{node{$\theta_{k}^{+--}$}
          child{node{$....$}}
          }
          child{node{$\theta_k^{+-+}$}}  
        }
        child {node{$\theta_k^{++}$}
          child{node{$\theta_{k}^{++-}$}
          child{node{$....$}}
          }
          child{node{$\theta_k^{+++}$}}  
        }
      };
  \end{tikzpicture}
  \caption{An illustration of our proposed internal split of imagenet classes.} \label{fig:weight_tree}
  \end{figure}
  
\begin{figure}[ht!]
  \centering
  \begin{subfigure}{0.2\textwidth}
      \includegraphics[width=\textwidth]{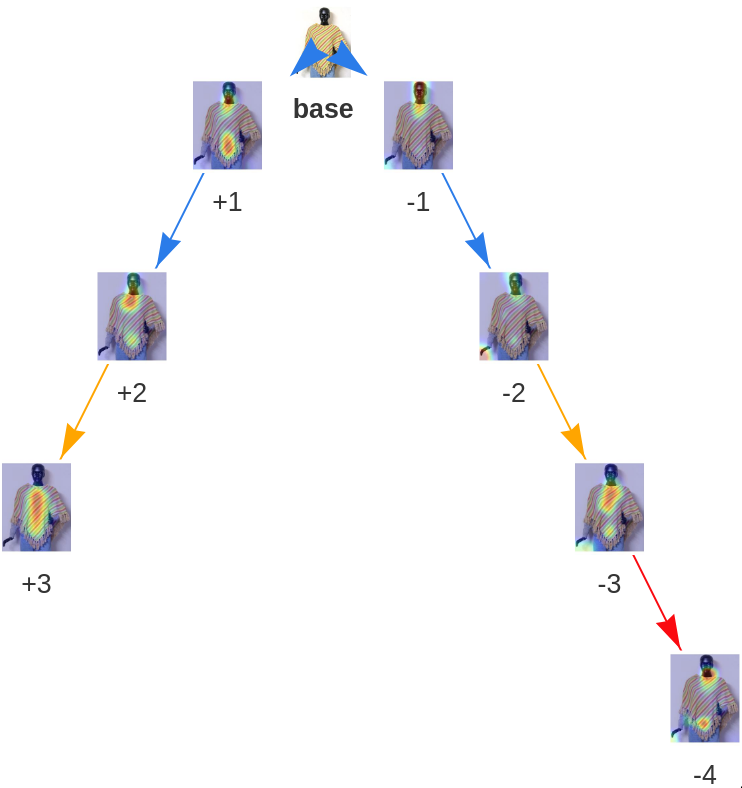}
      \caption{Target Imagenet class: \textit{poncho}. Notice that in the terminal node of positive branch only features related to poncho are activated whereas the face is activated in the negative branch.}
      % \label{fig:first}
  \end{subfigure}
  \hspace{0.2mm}
  \begin{subfigure}{0.2\textwidth}
      \includegraphics[width=\textwidth]{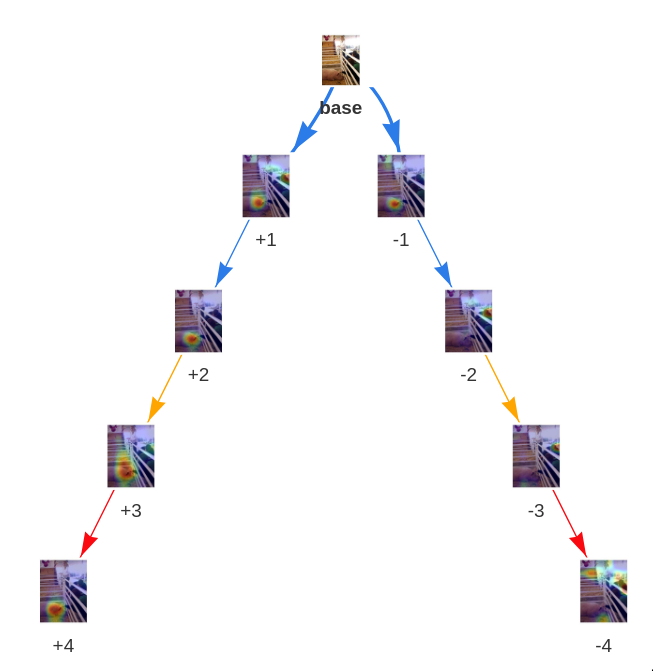}
      \caption{Target Imagenet class: \textit{hog}. Portions of the fence are activated in the negative branch whereas only the hog is activated in positive branch.}
  \end{subfigure}
  \hspace{0.2mm}
  \begin{subfigure}{0.2\textwidth}
      \includegraphics[width=\textwidth]{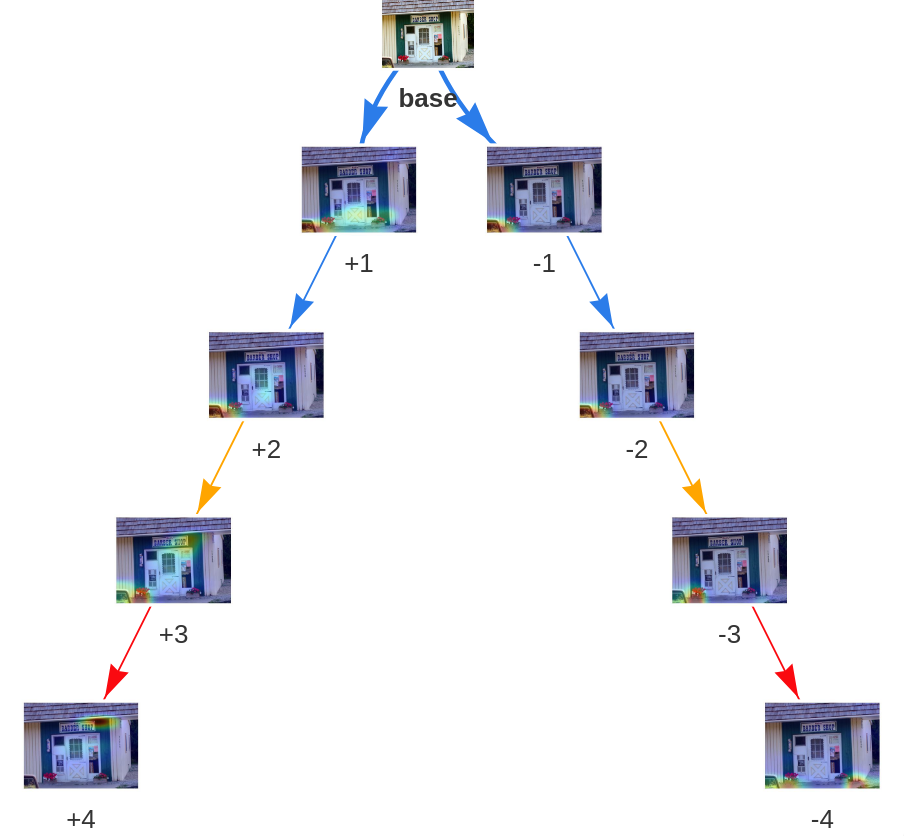}
      \caption{Target Imagenet class: \textit{Barber Shop}. The text associated with barbershop are activated in the positive branch.}
  \end{subfigure}
          
  \caption{Smooth Grad Cam++ \ref{fig:sm_viz} visualization of activated features. In the positive branch we only use the subset of neurons in the final fully connected layer identified
  using our proposed method \ref{fig:weight_tree}. According to our interpretation, the positive branch should yield the most robust features for the class whereas the the negative branch should identify the
  most confusing features.}
  \label{fig:sm_viz}
\end{figure}

While fitting ALD distributions to the FC layer weights for each class $\theta_k$, we observed a recursive pattern in the isolated weights. Let $\theta_k^+$ denote
the weights for the positive sub-class (i.e. the possibility that the sample belong to class k) \& $\theta_k^-$ denote the weights for the negative class (i.e. the possibility that the sample is outside training set).
We hypothesize that $\theta_k^+$ and $\theta_k^-$ can be further sub-divided and fitted with ALD. An illustration of our idea is given in figure \ref{fig:weight_tree}.
The proposed recursive partition of weights $\theta_k$ can help us isolate the most discriminative and most confusing feature according of the input image for the neural network. More specifically,
let $\theta_k^{+++}$ denote the final tree node in the positive part after the proposed split of the weights $\theta_k$. Since these tree nodes were always encountered while
maximizing the possibility of kth class, the associated neurons should activate on the most vital feature for the kth class. Similarly the terminal weights in the negative branch $\theta_k^{---}$ should
encode the most confusing aspect for the kth class.\\
To verify our hypothesis, we used Smooth Grad-Cam ++\cite{omeiza2019smooth} to highlight image regions associated with the most activated positive $\theta_k^{++..}$  and negative terminal $\theta_k^{--..}$ weights for the target class.
\textbf{We strongly recommend the reader to download the plots from \href{https://drive.google.com/drive/folders/1aWOlXt20iZJGgaXFMusCTYmJkOyLImLj?usp=sharing}{Google Drive Link} for more detailed viewing.}
A visualization of the weights can be found in figure \ref{fig:sm_viz}. We applied our split method to identify the most important neurons in the final
fully connected layer in Resnet34 \cite{he2016identity}. The activation associated with only these neurons were used for the visualization. In the figure, we report activations
for each stage of the split for reference. For example: +2 \& -2 refers to the positive and negative branches after the first split. We used this notation to avoid clutter. A more descriptive notation could be $\theta_k^{++}$ and  $\theta_k^{--}$ as
used earlier in our previous illustration \ref{fig:weight_tree}.\\
In our experiments, we observed that it is difficult to interpret the activations of neurons from the 
intermediate stages. We found the activations in the terminal node easier for subjective interpretation. The visualizations were generated using images from validation set of Imagenet.

\section{Importance of explainability and prior attempts}
Traditional methods like Decision Trees and Rule based methods have high explanatory power but lag behind deep neural networks in performance. In recent years deep neural networks have become very easy to deploy which has led to their widespread usage. 
Hence the issue of interpretability and explainability has become even more pressing. Explainability is particularly important in high stake applications
like autonomous driving and medical diagnosis. Aside from the practical benefit of explainable decisions, interpretable deep neural networks help us
in distilling out the important aspects of network architecture and training methodology. These insights will help us to develop a new class of explainable neural networks which will
hopefully be more data efficient, have smaller number of neurons and be able to efficiently utilize inductive priors.\\
Currently popular neural networks are vastly over-parameterized, sometimes requiring millions of parameters to solve 
seemingly simple visual classification problems. This is a commonly accepted phenomenon but it is not intuitive. The natural 
world already has organisms like nematodes which display interesting behavioral patterns even though they have 
a very small number of neurons. The adaptability and robustness of such simple organisms in challenging environments points towards a gap in our understanding of neural networks.

Given the importance of explanability, many methods have been proposed to understand neural network predictions. Attribution based methods \& 
perturbation based methods are two most popular approaches. Attribution based methods aim at characterizing the response of 
neural networks by finding which parts of the network’s input are the most responsible for determining its output. These methods generally use 
backpropagation to track information from the network’s output back to its input, or an intermediate layer. Methods like GradCAM \cite{selvaraju2017grad} and Guided Backprop
are the most famous example of these kind of methods.
Approaches like RISE \cite{petsiuk2018rise} and Meaningful Perturbations \cite{fong2017interpretable} belong to the perturbation family of methods. These methods perturb the inputs to the model and observe resultant changes to the output. Although these methods are interesting, they still have drawbacks. For example, methods like GradCAM may capture average network properties
but may not be able to characterize intermediate activations, or sometimes the model parameters.

\section{Conclusion and Future work}
In this work we have attempted to establish properties of the final fully connected layer weights in pre-trained image classification models. We have provided theoretical intuition
and experimental results for our claims. A more rigorous theoretical analysis may help uncover additional insights regarding the distribution of weights.\\
Following are few potential future work:
\begin{itemize}
  \item Faster model convergence \& model robustness on test data\\
  If there is prior knowledge about the data distribution then it can be incorporated during the supervised learning stage by adjusting the policy gradient during loss calculation step. This process can be particularly useful for imbalanced datasets.
  \item Pruning the final fully connected layer\\
  If a prior distribution is known for the dataset then network weights which show high deviation from the prior distribution can be pruned away without hurting the model accuracy significantly.
\end{itemize}

\newpage
%----------------------------------------------------------------------------------------
%	REFERENCE LIST
%----------------------------------------------------------------------------------------

\bibliography{ref} 
\bibliographystyle{ieeetr}

%----------------------------------------------------------------------------------------

\end{document}

%% file: defs.tex
\newcommand{\reward}{r}
\newcommand{\policy}{\pi}

\newcommand{\traj}{\tau}

\newcommand{\bs}{\mathbf{s}}
\newcommand{\ba}{\mathbf{a}}

\newcommand{\E}{\mathbb{E}}